\documentclass{article} 
\usepackage{iclr2025_conference,times}
\usepackage{graphicx}
\usepackage{subcaption}

\usepackage{amsmath,amsfonts,bm}

\def\eqref#1{equation~\ref{#1}}

\def\1{\bm{1}}

\DeclareMathAlphabet{\mathsfit}{\encodingdefault}{\sfdefault}{m}{sl}
\SetMathAlphabet{\mathsfit}{bold}{\encodingdefault}{\sfdefault}{bx}{n}

\DeclareMathOperator*{\argmax}{arg\,max}

\usepackage{hyperref}
\usepackage{url}
\usepackage{booktabs}

\title{Domain-Invariant Prompt Learning\\for Vision-Language Models}

\author{Arsham Gholamzadeh~Khoee, Yinan Yu \& Robert Feldt \\
Department of Computer Science and Engineering\\
Chalmers University of Technology\\
Gothenburg, Sweden \\
\texttt{\{khoee,yinan,robert.feldt\}@chalmers.se} \\
}

\iclrfinalcopy 
\begin{document}

\maketitle

\begin{abstract}

Large pre-trained vision-language models like CLIP have transformed computer vision by aligning images and text in a shared feature space, enabling robust zero-shot transfer via prompting. Soft-prompting, such as Context Optimization (CoOp), effectively adapts these models for downstream recognition tasks by learning a set of context vectors. However, CoOp lacks explicit mechanisms for handling domain shifts across unseen distributions. 
To address this, we propose Domain-invariant Context Optimization (DiCoOp), an extension of CoOp optimized for domain generalization. By employing an adversarial training approach, DiCoOp forces the model to learn domain-invariant prompts while preserving discriminative power for classification. Experimental results show that DiCoOp consistently surpasses CoOp in domain generalization tasks across diverse visual domains.

\end{abstract}

\section{Introduction}
\label{sec:intro}
The emergence of large language models (LLMs) has demonstrated their remarkable capabilities, which are now widely recognized. Building upon this success, vision-language models have emerged as a powerful alternative for visual representation learning. These models aim to align images and their corresponding raw text using two distinct encoders: one for text and the other for vision. For instance, CLIP~\citep{radford2021learning}, one of the most prominent vision-language models, uses contrastive learning to pull together images and their textual descriptions while pushing apart unmatched pairs in the feature space. Unlike traditional vision models, which are pre-trained on fixed sets of discrete class labels using cross-entropy loss, vision-language models leverage textual semantics for training, allowing them to better understand textual information~\citep{yang2024embodied}. By pre-training on large-scale datasets, these models can learn diverse visual concepts and transfer them effectively to downstream tasks through prompting. For example, in image classification tasks, task-relevant sentences describing categories can be fed to the text encoder, and the resulting text features can be compared with image features produced by the image encoder.

Several studies have highlighted the importance and nuances of prompts for achieving optimal performance on downstream datasets. \citet{zhou2022learning} proposed Context Optimization (CoOp), a novel approach for finding optimized prompts in image classification tasks. In particular, CoOp transforms prompt engineering from a manual process into an optimization problem by using some learnable numerical vectors called \emph{context vectors}. 
CoOp has been shown to outperform handcrafted prompts and exhibits stronger robustness than standard zero-shot models with manually designed prompts. 

However, while CoOp demonstrates some resilience to domain shifts, it does not explicitly address the challenge of domain invariance in prompt learning to handle distribution shifts or unseen domains. These challenges are formulated as domain adaptation (DA) and domain generalization (DG) in the literature. DA focuses on adapting models from source to target domains with access to target domain data during training. In contrast, DG aims to generalize to unseen domains without such access~\citep{zhou2022domain}. DG is particularly relevant in real-world scenarios, where models are trained on specific datasets but must perform well on new, previously unseen data distributions~\citep{khoee2024domain}. 

To achieve both high accuracy for the task and robustness to domain shift, we aim to design a prompt that is highly effective for class discrimination but incapable of identifying the domain of the input data. This idea aligns with the definition of a good cross-domain representation proposed by~\citet{ben2010theory}, which emphasizes that a model should prevent domain distinction while maintaining class discrimination. In other words, the model should emphasize task-relevant information while promoting domain confusion to achieve effective generalization across domains.

In this work, we propose Domain-invariant Context Optimization (DiCoOp), an extension of CoOp designed specifically for domain generalization tasks. DiCoOp applies adversarial training principles to prompt learning, explicitly promoting domain invariance within the learnable context vectors.
We introduce three implementations of DiCoOp to explore different prompt structures: (1) Domain-First Prompting (DFP), which separates domain and class tokens, placing domain tokens first; (2) Class-First Prompting (CFP), similar to DFP but with class tokens placed before domain tokens; and (3) Shared Context Prompting (SCP), which does not explicitly separate domain and class tokens, instead using a shared context for joint learning.

As a summary, we have contributed the following:
\begin{itemize}
    \item We introduce DiCoOp, an extension of CoOp that leverages domain adversarial prompt learning using the Gradient Reversal Layer (GRL) to enhance the robustness of VLMs against domain shifts effectively.
    \item We propose three distinct prompting strategies—SCP, DFP, and CFP—to explore how prompt design affects domain generalization in vision-language models. Notably, DFP and CFP systematically split and freeze domain- and class-specific tokens, preserving class-discriminative knowledge while addressing domain invariance.
    \item DiCoOp outperforms its baseline, which is CoOp, on PACS (using ResNet-50) and Mini-DomainNet (using ViT-B/16) datasets, demonstrating the robustness of DiCoOp across domain generalization tasks.
\end{itemize}

\section{Proposed Method}
\label{sec:method}
Let $\mathcal{D}^s = \{\mathcal{D}_i^s\}_{i=1}^n$ denote a set of $n$ source domains, each containing input data $x_i \in \mathcal{X}_i$ and corresponding labels $y_i \in \mathcal{Y}$. The probability distribution of each source domain, denoted as $P(\mathcal{D}_i^s)$, differs across domains such that $P(\mathcal{D}_i^s) \neq P(\mathcal{D}_j^s)$ for all $i,j \in {1,\dots,n}$ where $i \neq j$. In DG, our goal is to train a model on these source domains that generalizes well to an unseen target domain $\mathcal{D}^t$, where the target domain distribution $P(\mathcal{D}^t)$ is distinct from all source domain distributions, i.e., $P(\mathcal{D}^t) \neq P(\mathcal{D}_i^s)$ for all $i \in {1,\dots,n}$. This setup is commonly referred to as the multi-source domain generalization problem~\citep{khoee2024domain}.

Due to page constraints, we refer to the related work and preliminaries in Appendices \ref{sec:related} and \ref{sec:preliminaries}, respectively, which provide a detailed description of the Vision-Language model CLIP \citep{radford2021learning} and the learnable soft-prompting method CoOp~\citep{zhou2022learning}.

\subsection{Domain-invariant Context Optimization}
\begin{figure*}[ht]
\centering
\includegraphics[width=1\linewidth]{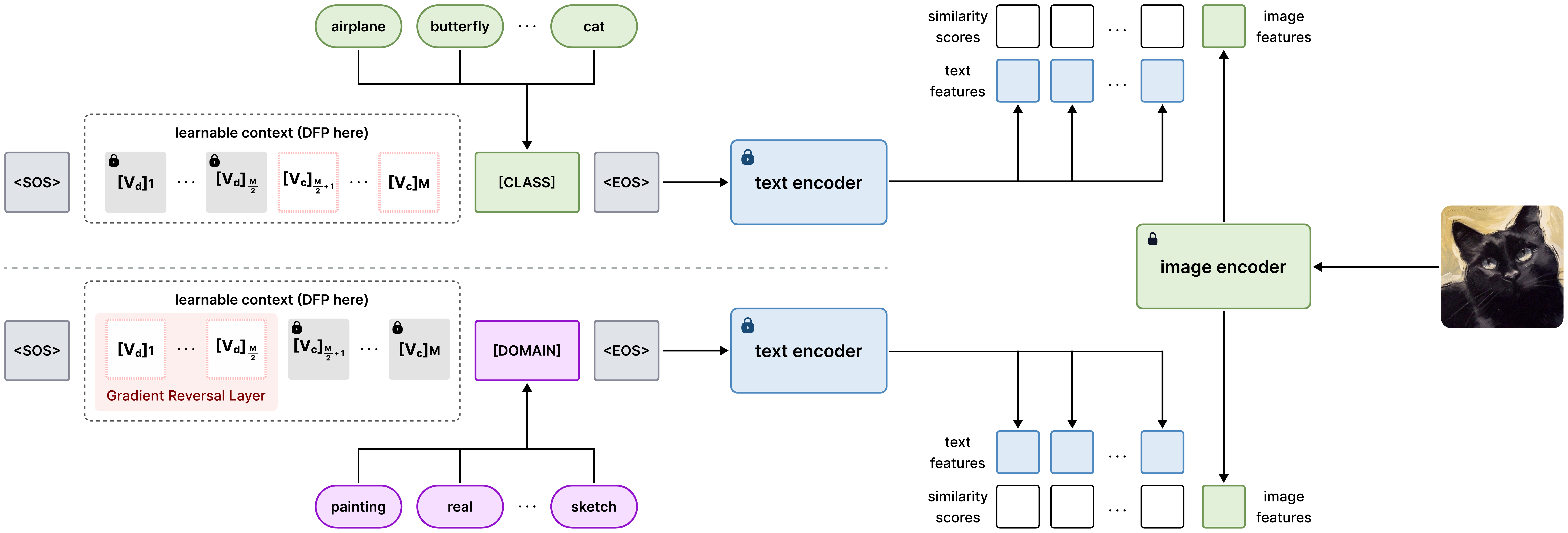} 
\caption{Overview of Domain-invariant Context Optimization (DiCoOp). Domain First Prompting (DFP) is illustrated, where the first half of the prompt is dedicated to domain information, and the remaining half is dedicated to class information. During domain-related optimization, the class tokens remain frozen, and vice versa.}
\label{fig:architecture}
\end{figure*}

Our objective is to learn domain-invariant prompts that reduce domain bias and enable robust performance on unseen domains. Drawing inspiration from Domain Adversarial Neural Networks (DANN)~\citep{ganin2016domain}, we incorporate a Gradient Reversal Layer (GRL) into the prompt tuning process.  Our approach exploits both class names and (source) domain names: we perform standard prompt tuning for classification while applying adversarial training (via GRL) to encourage domain generalization by maximizing domain-specific feature distinguishability. We aim to optimize the context vectors to maintain strong class discrimination while ensuring invariance to domain differences. An overview of this architecture is shown in Figure~\ref{fig:architecture}.

We optimize learnable context vectors $\mathbf{v}$ by minimizing the negative log-likelihood of the ground-truth label of classes and maximizing the negative log-likelihood of the ground-truth label of domains. We initially assume a shared set of learnable context vectors $\mathbf{v}$ that simultaneously undergo standard gradients (for class prediction) and reversed gradients (for domain prediction). Alternatively, these learnable vectors can be split into two parts: one dedicated to category classification information and another for domain-invariant information.
As a result, the overall training objective combines classification \emph{(cls)} and domain adversarial \emph{(dom)} losses:
\begin{equation}
    \begin{split}
    \mathcal{L}(\mathbf{v}) = \mathcal{L}_{cls}(\mathbf{v} ) - \lambda\mathcal{L}_{dom}(\mathbf{v} ) 
    = -\sum_{i} y_i^c log P(i | x) + \lambda\sum_{j} y_j^d log P(j | x),
    \end{split}
\end{equation}
where $y^c$ and $y^d$ are one-hot encodings of the ground-truth class and domain labels, respectively, and $\lambda \geq 0$ controls the strength of domain adversarial training. $P(i|x)$ in the classification loss is the probability that the input image $x$ belongs to the $i-$th class, while $P(j|x)$ in the domain adversarial loss is the probability that $x$ comes from the $j-$th domain.

The GRL is essential for domain-invariant context optimization (DiCoOp). During the forward pass, GRL acts as an identity function, allowing standard computation of both class and domain predictions. However, during backpropagation, the GRL multiplies the gradient by $- \lambda$ for the domain-specific portion of the context. This adversarial approach encourages domain context vectors to become domain-invariant while preserving class discrimination. 

Although we describe the method with a shared context vectors $\mathbf{v}$, referred to as \textbf{Shared Context Prompting (SCP)}, the tokens can be split into domain and class segments:

\textbf{Domain-First Prompting (DFP):} 
    The first half of the prompt is designated for domain-specific tokens, and the second half for class-specific tokens, i.e.,
    \[
    \mathbf{v} 
    = [\mathbf{v}_d]_1 \cdots [\mathbf{v}_d]_{\tfrac{M}{2}} 
      \;[\mathbf{v}_c]_{\tfrac{M}{2}+1} \cdots [\mathbf{v}_c]_{M}.
    \]
    During domain-related optimization, only the domain-specific tokens are updated (class-specific tokens remain frozen), and vice versa. This explicit separation helps maintain clear boundaries between domain and class information. 

\textbf{Class-First Prompting (CFP):}
    Similar to DFP, but reversed: class-specific tokens come first, followed by domain-specific tokens, i.e.,
    \[
    \mathbf{v} 
    = [\mathbf{v}_c]_1 \cdots [\mathbf{v}_c]_{\tfrac{M}{2}} 
      \;[\mathbf{v}_d]_{\tfrac{M}{2}+1} \cdots [\mathbf{v}_d]_{M}.
    \]
    Class-specific tokens are frozen during domain-related optimization, and domain-specific tokens are frozen during class-related optimization. Like DFP, CFP preserves a strict separation 
    between domain and class segments. 

\subsection{Training and Inference}
During \emph{training}, we learn domain-invariant context vectors by performing two forward passes for each input image $x$:

\textbf{1. Class pass:} We form the prompt for class $k$ by concatenating the learnable context $\mathbf{v}$ with the class token $[CLASS]_k$, i.e.:
\begin{equation}
    \begin{split}
    t_k^c = concat(\mathbf{v}, [CLASS]_k).
    \end{split} 
\end{equation} 
This prompt is fed into the model for classification, and standard gradients update $\mathbf{v}$ to minimize class prediction loss.

\textbf{2. Domain pass:} We form the prompt for domain $p$ by concatenating the same context $\mathbf{v}$ with the domain token $[DOMAIN]_p$, i.e.:
\begin{equation}
    \begin{split}
    t_p^d = concat(\mathbf{v}, [DOMAIN]_p).
    \end{split} 
\end{equation} 
This prompt is fed into the model for domain prediction through the GRL, so the gradients are reversed to maximize domain prediction loss, encouraging $\mathbf{v}$ to become domain-invariant.

By alternating between class and domain passes, we learn context vectors $\mathbf{v}$ that balance accurate class discrimination with minimal domain bias. 

At \emph{inference} time, we no longer have access to domain labels. We only perform the class pass, utilizing the learned domain-invariant context $\mathbf{v}$ to predict class labels for incoming images, ensuring robust classification across unseen domains.

\section{Experimental Results}
\label{sec:results}

\begin{figure}[ht]
     \centering
     \begin{subfigure}
     {0.49\textwidth}
         \centering
         \includegraphics[width=\textwidth]{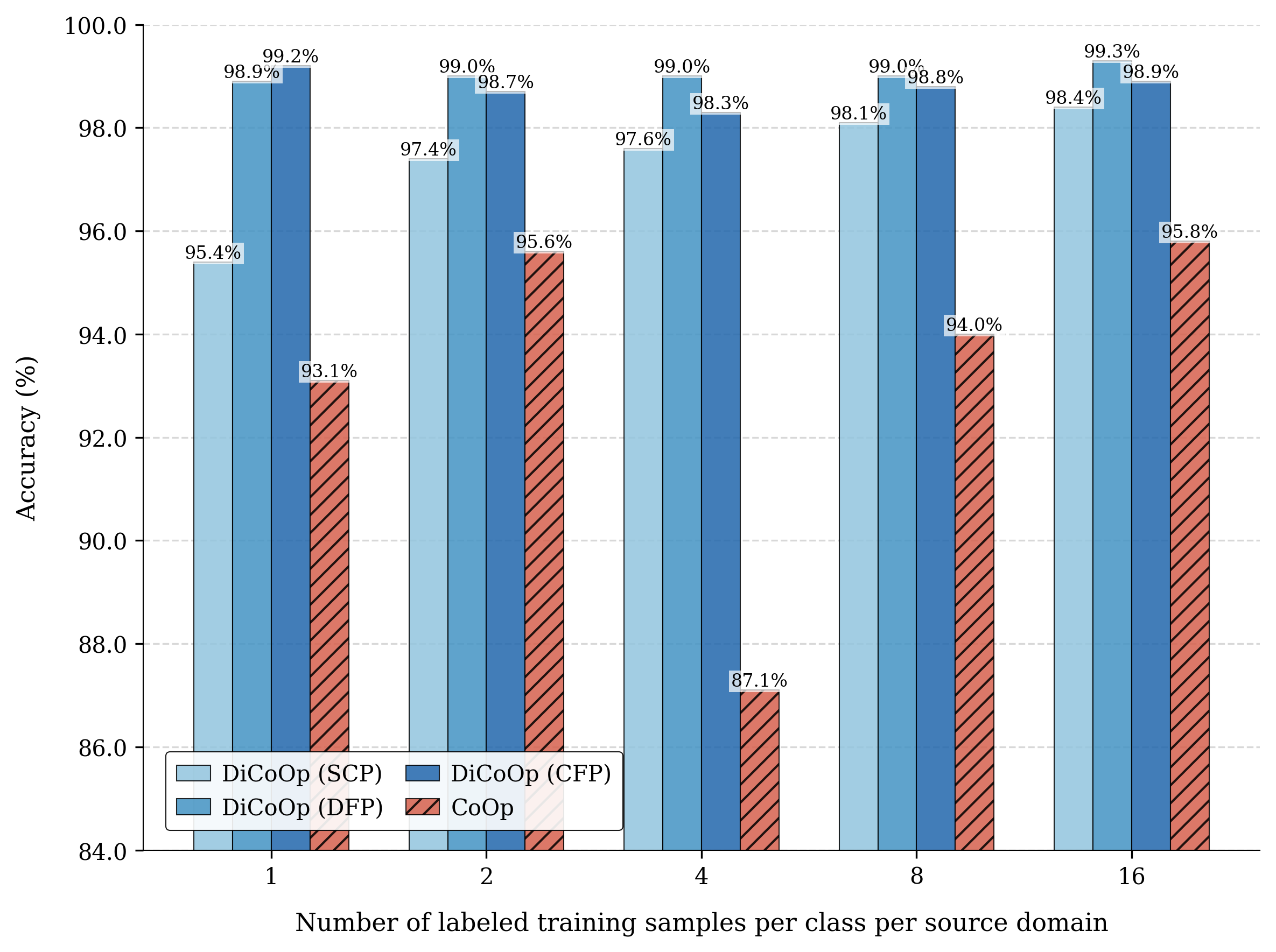}
         \caption{Photo}
         \vspace{0mm}
     \end{subfigure}
     \hfill
     \begin{subfigure}
     {0.49\textwidth}
         \centering
         \includegraphics[width=\textwidth]{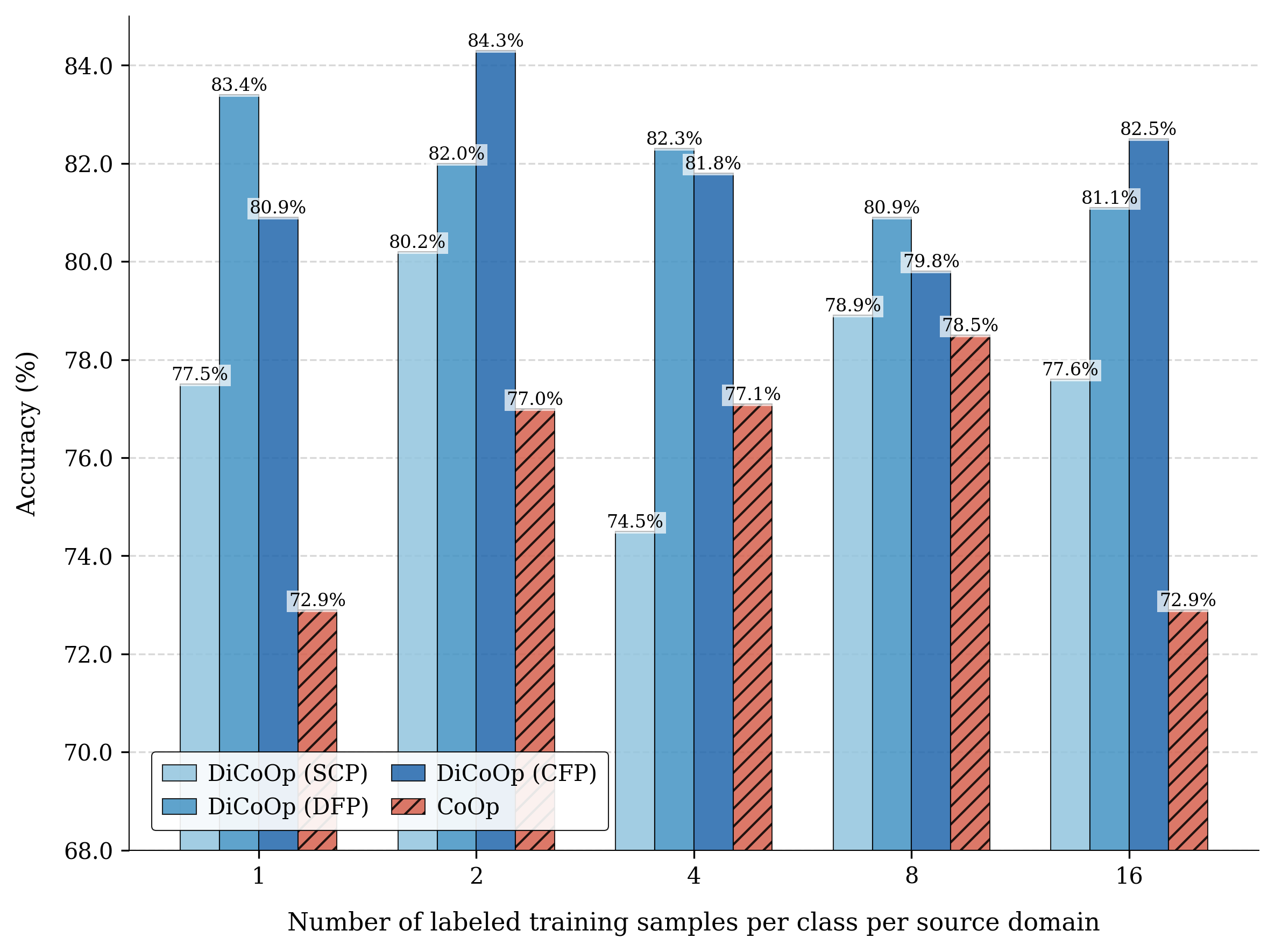}
         \caption{Sketch}
         \vspace{0mm}
     \end{subfigure}
     \begin{subfigure}
     {0.49\textwidth}
         \centering
         \includegraphics[width=\textwidth]{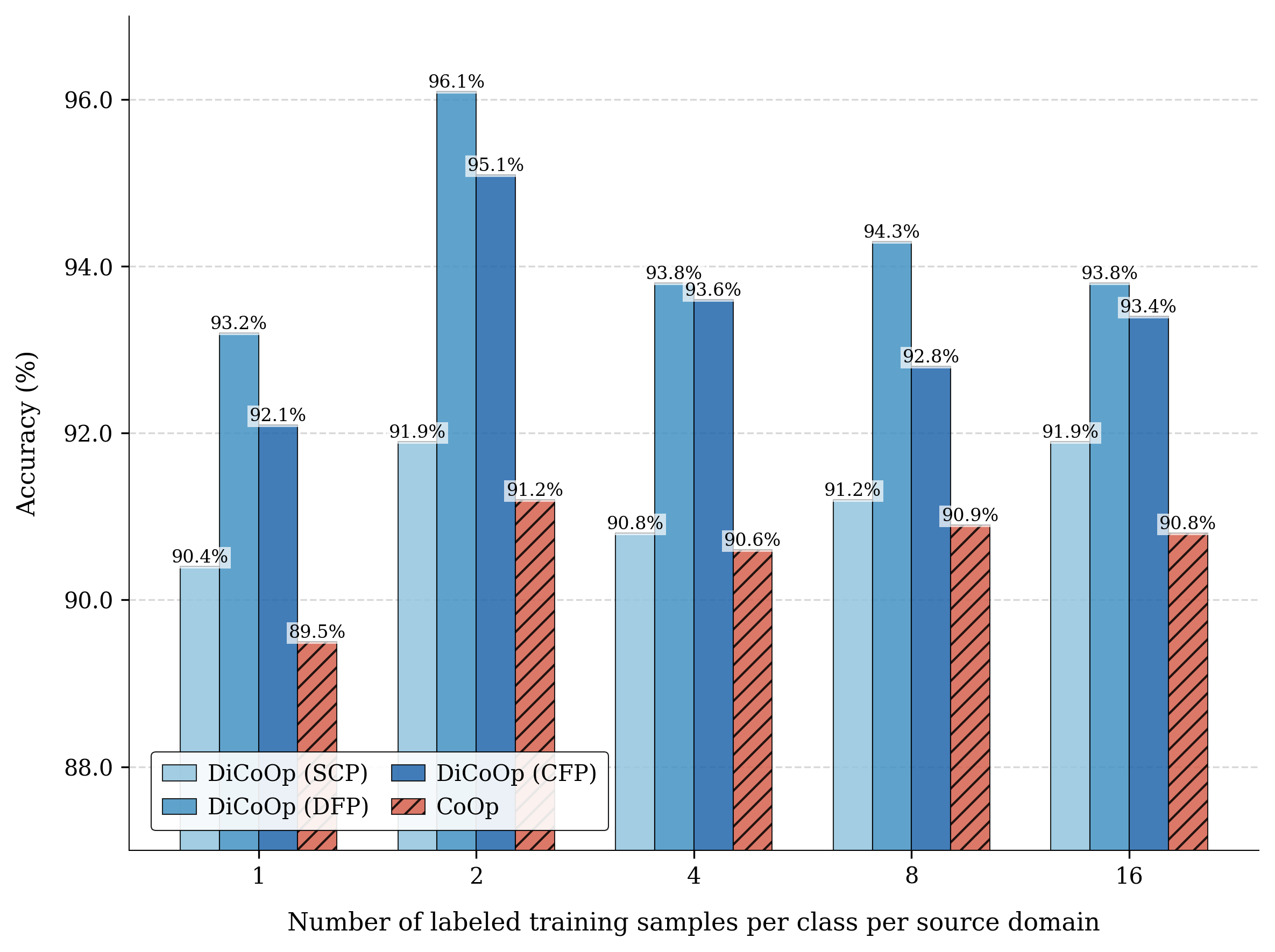}
         \caption{Cartoon}
     \end{subfigure}
     \hfill
     \begin{subfigure}
     {0.49\textwidth}
         \centering
         \includegraphics[width=\textwidth]{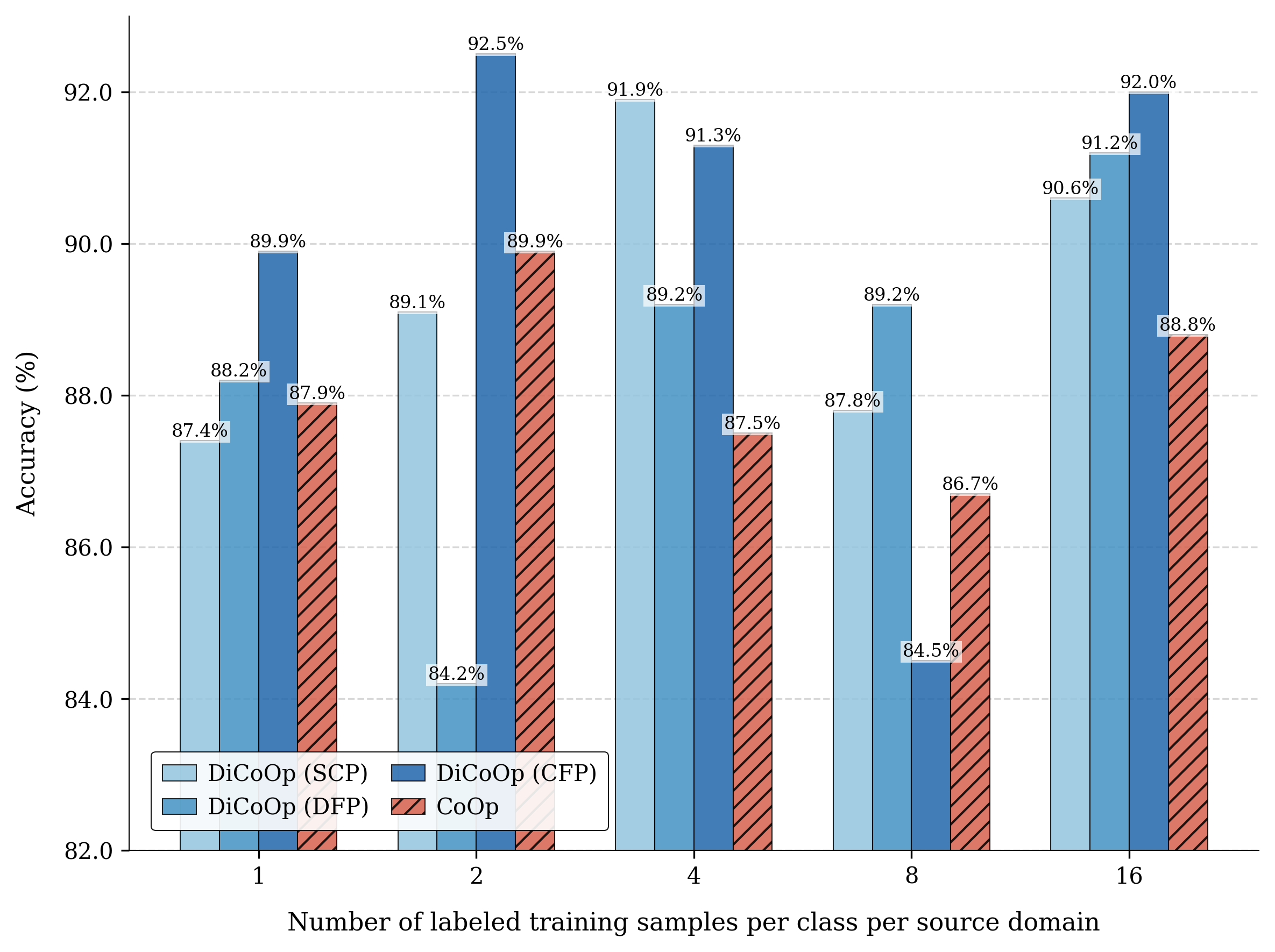}
         \caption{Art Painting}
     \end{subfigure}
        \caption{Results of few-shot learning on the PACS dataset using the leave-one-domain-out technique. Each plot is defined by the domain name that is left out during prompt learning, and testing is performed on that same domain.}
        \label{fig:plots}
\end{figure}

We evaluate our model on two publicly available datasets. i) PACS~\citep{li2017deeper}: This dataset spans four contrasting domains (Photo, Art Painting, Cartoon, and Sketch) and includes seven object categories. ii) Mini-DomainNet~\citep{yue2024less, tang2024source}: This dataset consists of four different domains (Clipart, Painting, Sketch, and Real), each containing images from 126 categories.

To assess domain generalization performance, we use a leave-one-domain-out strategy: one domain is held out as the target (test) domain, while the remaining domains serve as source domains for training. In all experiments, we set the prompt context length ($M$) to 16, following~\citet{zhou2022learning}.

We compare DiCoOp—with all three variants (SCP, DFP, CFP)—to CoOp as the baseline. For the PACS dataset, we use ResNet-50~\citep{he2016deep} as the backbone image encoder ($f_v$) and test 1-shot, 2-shot, 4-shot, 8-shot, and 16-shot settings. In these $n$-shot experiments, each source domain contributes $n$ labeled examples per class. To ensure a fair comparison, we evaluate the baseline under the same conditions.

Results on PACS are illustrated in Figure~\ref{fig:plots}, showing the classification accuracy for each target domain versus the number of labeled training examples per class per domain. Overall, DiCoOp demonstrates greater robustness and better generalization than CoOp. Among DiCoOp variants, CFP and DFP outperform SCP, suggesting explicit separation of domain/class segments improves handling of domain variation. CFP demonstrates the strongest cross-domain consistency, consistently achieving high (and often top) accuracy. Meanwhile, SCP shows less reliable performance, often yielding results comparable to or occasionally lower than the baseline CoOp, indicating that shared context vectors struggle to effectively disentangle domain and class information.

For Mini-DomainNet, we switch to a ViT-B/16~\citep{dosovitskiy2020image} backbone while using the 16-shot setting. Table~\ref{tab:mini} reports the accuracy on each target domain. DiCoOp outperforms CoOp on all target domains, underscoring the robustness of domain-invariant prompt tuning. Notably, DiCoOp (CFP) and DiCoOp (DFP) achieve the same average accuracy, improving the results by $2.23\%$ over CoOp, while DiCoOp (SCP) shows a $1.23\%$ improvement. These results affirm that the separation of domain and class tokens (DFP or CFP) enhances generalization compared to fully shared prompts (SCP).

\begin{table}[ht]
\centering
\renewcommand{\arraystretch}{1.2}
\setlength{\tabcolsep}{8pt}
\caption{Accuracy (\%) on Mini-DomainNet for domain generalization using a leave-one-domain-out approach. Each column shows results on the domain that has been left out, comparing domain generalization performance to the baseline. Bold numbers indicate the best accuracy in each column.}
\begin{tabular}{l c|c c c c|c}
\toprule[1.2pt]
\textbf{Methods} & \textbf{Backbone} & \textbf{Clipart} & \textbf{Painting} & \textbf{Sketch} & \textbf{Real} & \textbf{Avg.} \\
\midrule[0.4pt]
CoOp & ViT-B/16 & 83.5 & 80.3 & 76.6 & 88.7 & 82.27 \\
DiCoOp (SCP) & ViT-B/16 & 83.8 & 81.9 & 78.6 & 89.7 & 83.5 \\
DiCoOp (DFP) & ViT-B/16 & 83.9 & \textbf{83.2} & 79.8 & 91.1 & \textbf{84.5} \\
DiCoOp (CFP) & ViT-B/16 & \textbf{84.1} & 82.7 & \textbf{80.0} & \textbf{91.2} & \textbf{84.5} \\
\bottomrule[1.2pt]
\end{tabular}
\label{tab:mini}
\end{table}

\section{Conclusion}
\label{sec:conclusion}
Vision-language models have shown significant promise across various tasks. However, for specific downstream classification tasks, effectively generalizing these pre-trained models to unseen domains remains an open challenge. To bridge this gap, we introduced DiCoOP, a novel framework to improve domain generalization in vision–language models. Built upon CLIP, DiCoOP learns domain-invariant prompt tokens by incorporating a domain adversarial loss into prompt tuning. Specifically, we employ a Gradient Reversal Layer (GRL) to penalize domain classification, thereby mitigating domain bias using prompts and encouraging domain invariance. This study serves as an initial exploration of incorporating adversarial training into prompt learning to learn domain-invariant prompts and enable robust performance on unseen domains. Our experiments on two benchmark datasets demonstrate that DiCoOP outperforms its baseline (CoOp), highlighting the effectiveness of adversarial prompt tuning.
In future work, we plan to extend DiCoOP to more challenging domain generalization applications such as person re-identification and medical imaging to investigate its effectiveness and guide further advancements in domain generalization for emerging foundation models.

\subsubsection*{Acknowledgments}
The research was supported by two funding sources: the Wallenberg AI, Autonomous Systems and Software Program (WASP) funded by the Knut and Alice Wallenberg Foundation, and the Chalmers Gender Initiative for Excellence (Genie).

\bibliography{iclr2025_conference}
\bibliographystyle{iclr2025_conference}

\appendix
\section{Appendix}

\subsection{Related Works}
\label{sec:related}
Vision-language models (VLMs), such as CLIP~\citep{radford2021learning}, ALIGN~\citep{jia2021scaling}, and VisualBERT~\citep{li2019visualbert}, integrate visual and textual data to enhance multimodal understanding, achieving state-of-the-art performance across diverse computer vision tasks. A key advancement in leveraging these models is prompt learning, which adapts pre-trained VLMs to downstream tasks by optimizing task-specific text prompts. In this regard, CoOp~\citep{zhou2022learning} developed prompt tuning for few-shot image classification by learning continuous prompt vectors, establishing a foundation for subsequent methods. Building on this, CoCoOp~\citep{zhou2022conditional} introduced conditional prompts that dynamically adjust to input images, enhancing the generalization for image classification. ProGrad~\citep{zhu2023prompt} refined this further by selectively updating the prompt whose gradient is aligned (or non-conflicting) to the general knowledge to prevent prompt tuning from forgetting the general knowledge learned from VLMs. Taking a different approach, CLIP-Adapter~\citep{gao2024clip} focused on fine-tuning feature adapters in both visual and language branches to improve CLIP's classification capabilities.

For DA challenges, several approaches have emerged. DAPL~\citep{ge2023domain} introduced domain-specific prompt tuning, though its requirement for explicit domain information limits practical applications. AD-CLIP~\citep{singha2023ad} aimed to create domain-agnostic prompts through prompt learning; however, its reliance on distribution alignment poses challenges with limited target domain samples. Recent work has explored DG using CLIP by incorporating domain-specific learnable residuals in text embeddings alongside domain-shared residuals~\citep{feng2024rethinking}. The latter is then used at inference to capture common knowledge across domains. However, the effectiveness of these simple domain priors depends heavily on how precisely the domain can be described in natural language, given that prompts are handcrafted.

One notable approach to address domain shifts is the Domain-Adversarial Neural Networks (DANN) proposed by \citet{ganin2016domain}, which trains neural networks to be both discriminative (for the classification task) and domain-invariant. DANN minimizes the loss of the main label classifier while maximizing the loss of the domain classifier using Gradient Reversal Layers (GRL). GRL adversarially trains the network to confuse the domain classifier, encouraging the emergence of domain-invariant features during training. Inspired by DANN, our work integrates a GRL into prompt learning to address the challenges of handling domain shifts in prompt learning. 

\subsection{Preliminaries}
\label{sec:preliminaries}
We use CLIP as our backbone architecture, which consists of an image encoder $f_v(\cdot)$ (either ResNet~\citep{he2016deep} or ViT~\citep{dosovitskiy2020image}) and a text encoder $f_t(\cdot)$ (BERT~\citep{devlin2018bert}). These encoders project their respective inputs from high-dimensional spaces into a shared low-dimensional feature space.

CLIP is trained on image-text pairs using contrastive learning, where associated image-text pairs serve as positive samples and non-associated pairs as negative samples. The contrastive objective maximizes the similarity between positive pairs while minimizing the similarity between negative pairs, effectively aligning image and text representations in the same feature space.

For zero-shot classification, given an input image $x$ and a set of $K$ textual category descriptions, the probability that $x$ belongs to $i-$th category is computed as:
\begin{equation}
 P(i | x) = \frac{\exp(<f_t(t_i), f_v(x)>/\tau)}{\sum_{k=1}^K \exp(<f_t(t_k), f_v(x)>/\tau)},
 \end{equation}
where $\tau$ is the temperature hyperparameter and $<\cdot,\cdot>$ denotes cosine similarity. The predicted class $\hat{y}$ is then determined by:
\begin{equation}
 \hat{y} = \argmax_k P(k|x).
 \end{equation}

Traditionally, the input text consists of manually designed prompts composed of discrete tokens. These prompts are transformed into fixed vectors in the word embedding space. However, these fixed embeddings may be sub-optimal for category representation~\citep{ge2023domain}. To address this, we can optimize continuous embeddings of the prompt tokens, allowing for more precise semantic feature descriptions~\citep{lester2021power}. This is achieved through learnable context vectors $\mathbf{v}$, where the prompt for class $k$ is represented as:
\begin{equation}
\begin{split}
&\mathbf{v} = [\mathbf{v} ]_1[\mathbf{v} ]_2 \cdots [\mathbf{v} ]_M, \\
&t_k = concat(\mathbf{v}, [CLASS]_k),
 \end{split}
 \end{equation}
where each $[\mathbf{v}]_m$ ($m \in {1,2,\ldots,M}$) is a vector with the same dimension as the word embedding, and $M$ is the number of context tokens in the prompt. CoOp~\citep{zhou2022learning} optimizes these learnable context vectors by minimizing the negative log-likelihood of the ground-truth label:
\begin{equation}
 \mathcal{L}_{ce}(\mathbf{v}) = -\sum_i y_i \log P(i | x),
 \end{equation}
where $y$ represents the one-hot encoded ground-truth labels.

One key design consideration for this approach is determining the semantic meaning that each context vector $[\mathbf{v}]_m$ should capture, and defining an effective training strategy to optimize these context vectors accordingly.

\end{document}